\pdfoutput=1

\documentclass[11pt]{article}

\usepackage{ACL2023}

\usepackage{times}
\usepackage{latexsym}

\usepackage[T1]{fontenc}

\usepackage[utf8]{inputenc}
\usepackage{soul}
\usepackage{microtype}

\usepackage{inconsolata}
\usepackage{amsmath}
\usepackage{graphicx}

\usepackage{fancyhdr}
\title{Adapting LLMs to Hebrew: Unveiling DictaLM 2.0 with Enhanced Vocabulary and Instruction Capabilities}

\author{Shaltiel Shmidman\textsuperscript{1,†}, Avi Shmidman\textsuperscript{1,2,‡}, 
Amir DN Cohen\textsuperscript{2,†}, Moshe Koppel\textsuperscript{1,2,†} \\
\textsuperscript{1}DICTA / Jerusalem, Israel \\
\textsuperscript{2}Bar Ilan University / Ramat Gan, Israel \\ 
\texttt{\small \textsuperscript{†}\{shaltieltzion,moishk,amirdnc\}@gmail.com} \\
\texttt{\small \textsuperscript{‡}avi.shmidman@biu.ac.il}
}

\begin{document}
\maketitle
\thispagestyle{fancy}

\begin{abstract}
Training large language models (LLMs) in low-resource languages such as Hebrew poses unique challenges. In this paper, we introduce \texttt{DictaLM2.0} and \texttt{DictaLM2.0-Instruct}, two LLMs derived from the Mistral model, trained on a substantial corpus of approximately 200 billion tokens in both Hebrew and English. Adapting a pre-trained model to a new language involves specialized techniques that differ significantly from training a model from scratch or further training existing models on well-resourced languages such as English. We outline these novel training methodologies, which facilitate effective learning and adaptation to the linguistic properties of Hebrew. Additionally, we fine-tuned \texttt{DictaLM2.0-Instruct} on a comprehensive instruct dataset to enhance its performance on task-specific instructions. To rigorously evaluate our models, we introduce a new benchmark suite for Hebrew LLM evaluation, covering a diverse set of tasks including Question Answering, Sentiment Analysis, Winograd Schema Challenge, Translation, and Summarization. Our work not only addresses the intricacies of training LLMs in low-resource languages but also proposes a framework that can be leveraged for adapting other LLMs to various non-English languages, contributing to the broader field of multilingual NLP.

\end{abstract}

\section{Introduction}

The development of generative language models has significantly advanced natural language processing (NLP), enhancing the sophistication and contextual understanding in human-computer interactions \cite{openai2024gpt4technicalreport,geminiteam2024geminifamilyhighlycapable,claude2024model}. Leading state-of-the-art open-weight generative LLM models, such as Gemma and Llama \cite{gemma_2024,llama3modelcard}, are trained on trillions of tokens, but only a very small percentage of that data represents low-resource languages such as Hebrew \cite{touvron2023llama2}. Consequently, these models significantly underperform in such languages.

Training large language models (LLMs) in low-resource languages presents unique challenges, stemming from limited data availability, complex morphological structures, and the lack of robust evaluation frameworks tailored to these languages. Hebrew, with its rich morphology and limited large-scale corpora, exemplifies these difficulties. The tokenization process is particularly problematic, as it is not optimized for low resource languages like Hebrew, leading to inefficient compression ratios and suboptimal model performance \cite{petrov2023languagemodeltokenizersintroduce}.

In response to these challenges, we introduce \texttt{DictaLM2.0} and \texttt{DictaLM2.0-Instruct}, two generative language models specifically optimized for Hebrew. Built upon the Mistral model, these versions were trained on approximately 100 billion tokens each of Hebrew and English data. Adapting pre-trained models to a low-resource language like Hebrew involves specialized techniques that differ significantly from training models from scratch or further training existing models on well-resourced languages such as English. These include extending the tokenizer with Hebrew-specific tokens and performing embedding distillation to ensure effective learning and adaptation.

Additionally, we fine-tuned \texttt{DictaLM2.0-Instruct} on an instruct dataset to enhance its performance on task-specific instructions. This fine-tuning process is crucial for improving the model's ability to understand and execute task-specific commands accurately.

To address the evaluation gap in Hebrew NLP, we introduce a new benchmark suite for assessing Hebrew language models. This suite includes tasks such as Question Answering, Sentiment Analysis, Winograd Schema Challenge, Translation, and Summarization. Our comprehensive evaluation demonstrates that \texttt{DictaLM2.0} and \texttt{DictaLM2.0-Instruct} achieve state-of-the-art performance on these tasks, thereby setting new standards for Hebrew NLP.

The methodologies and evaluation frameworks presented in this work provide insights and potential pathways for adapting other LLMs to various non-English languages. This research contributes to the broader field of multilingual NLP by addressing the unique challenges posed by low-resource languages and offering scalable solutions for their integration into advanced language models.

\section{Related Work}

Interest in the community for adapting existing generative LLMs to languages other than English has been increasing. The first notable work is Chinese-LLaMA \cite{cui2024efficient}, which extended the LLaMA model \cite{touvron2023llama} and tailored it for Chinese. In that work, they extended the tokenizer of the model to include an additional 20,000 Chinese tokens, significantly improving the compression ratio, and subsequently continued pretraining the model on a large amount of Chinese text. Similar work was done by \citet{rakutengroup2024rakutenai7b}, extending LLaMA 2 to Japanese. However, as shown by \citet{zhu2023extrapolating}, training on cross-lingual translation data produces better performance compared to training on a monolingual corpus. The performance drop in Chinese-LLaMA seems to stem from the vocabulary extension, which requires a large monolingual corpus to train.

A similar approach was taken by \citet{basile2023llamantino}, who attempted to adapt the LLaMA 2 models \cite{touvron2023llama2} to Italian, leaving the tokenizer as is. As they note, the model demonstrated impressive reasoning in Italian, albeit the fluency and language skills were not up to par with the English models.

\citet{csaki2023efficiently} attempted a different approach when adapting language models to new languages. Instead of extending the tokenizer, they replaced infrequent tokens with new tokens tailored for the target language. In contrast with Chinese-LLaMA, they only replaced around 5,000 tokens, as they saw that replacing more tokens resulted in model quality degradation.

Overall, there have been many attempts to adapt language models to new languages. In this work, we employ a hybrid method, incorporating lessons learned from these previous efforts.

\section{Pre-training Data}
\label{sec:pretraining-data}

Our pre-training corpus consisted of 50\% Hebrew and 50\% English data. The English data was sourced from the Slim-Pajama corpus \cite{cerebras2023slimpajama}, sampled to match the quantity of Hebrew data available. In total, our Hebrew corpus comprises 35 billion words, the largest collection of Hebrew texts known to us.

This corpus translates into 95 billion tokens using our extended tokenizer (see Section \ref{sec:model-setup}). For comparison, the same corpus translates into approximately 204 billion tokens using the Mistral tokenizer \cite{jiang2023mistral}, or 196 billion tokens using the GPT-4 tokenizer.

Inspired by the works of \citet{zhu2023extrapolating,wendler2024llamas}, we also included a corpus of aligned translation sentence pairs to help the model map between its internal English knowledge and Hebrew. We used the CCMatrix corpus of Hebrew-English pairs \cite{schwenk2020ccmatrix} and converted each pair into text using set templates. An example template is: \texttt{"The translation of \{X\} in Hebrew: \{Y\}"}, where \texttt{X} is the sentence in English, and \texttt{Y} is the sentence in Hebrew.

\subsection{Hebrew Data}

\subsubsection{Sources}

The Hebrew data was collected from a wide range of sources, including available open-source corpora, internal web scraping, Hebrew books scanned and digitized in-house, and partnerships with companies that graciously provided their internal data for training. The data can be broken down into four main sources:

\begin{itemize}

\item \textbf{Internet}: Approximately 50\% of the data. This category includes web crawls such as C4 \cite{2019t5}, OSCAR \cite{OrtizSuarezSagotRomary2019} and Wikipedia. We processed over 150TB of unprocessed crawl data when extracting this data. 

\item \textbf{Social Media}: Approximately 28\% of the data. This data includes various social media corpora such as Hebrew twitter posts and online Hebrew blogs. 

\item \textbf{News}: Approximately 16\% of the data. This includes human transcriptions of TV and Radio, as well as various news sites.

\item \textbf{Other}: Approximately 6\% of the data. This includes various small corpora such as the Ben Yehuda project \cite{benyehuda_2024}, the Sefaria project \cite{sefaria_2024}, and additional Hebrew books digitized in-house. 

\end{itemize}

\subsubsection{Cleaning \& Filtering}

Our entire Hebrew corpus underwent a rigorous cleaning and filtering process to eliminate gibberish, low-quality, and irrelevant data, and to censor private information.

\textbf{Cleaning}: We began by replacing all sequences of content from languages other than Hebrew or English with a designated \texttt{<foreign>} token, ensuring that our corpus contains purely Hebrew and English text. This step was particularly important for corpora like Wikipedia, where entries often include foreign language text in parentheses, especially at the beginning of articles.

Next, we removed text markers such as XML/HTML tags, control characters (e.g., LTR mark, non-breaking space), etc. Finally, we replaced phone numbers, URLs, and emails with designated tokens (e.g., \texttt{<url>}), and removed duplicate paragraphs, keeping only the first instance if the same paragraph repeated more than twice in a row.

\textbf{Filtering}: We apply two levels of filtering to each document. The first level filters out documents which do not meet a specified threshold in terms of the percentage of Hebrew characters. The second level involves statistical checks which verify that the text within the document reflects the normative expectations of a Hebrew document; for instance, we constructed a histogram of all the Hebrew words in a gold corpus of ~1B words, and then for each document we checked to make sure that a majority of the words in the new document were frequent in the gold corpus, thus ensuring that the content contained real Hebrew words. The filtering thresholds were adjusted for each data source, with a human expert reviewing a sample of the filtered data to confirm that the thresholds were set correctly.

\subsubsection{De-duplication}

We ran our entire Hebrew corpus through a deduplication process to remove any document duplicates. We used the MinHash algorithm \cite{broder1997resemblance} to compute document similarities, employing an LSH index \cite{indyk1998approximate} for fast lookups. We set a threshold of 0.8, meaning two documents are considered duplicates if they have an estimated Jaccard similarity of 80\% or higher.

\section{Model Setup}
\label{sec:model-setup}

\subsection{Base Model}

Leading state-of-the-art generative LLM models are trained with trillions of tokens \cite{gemma_2024,llama3modelcard}, making their reproduction very expensive. Therefore, we initialize our model from an existing SOTA model and continue pretraining it, leveraging the vast amount of data the model was already trained on. We chose to initialize our model from the Mistral-7B-v0.1 checkpoint \cite{jiang2023mistral}, which was the leading SOTA open-source model at the time we began our experiments.

\subsection{Tokenizer Extension}

One major pitfall of the existing open-source SOTA models is the tokenization process. All modern generative LLMs use a tokenizer based on the BPE algorithm \cite{sennrich-etal-2016-neural}, which is a statistical method that creates tokens based on frequency in a corpus. Due to the scarcity of Hebrew in the training corpora of these models, there are very few Hebrew tokens in their vocabularies, resulting in a very inefficient compression ratio. This inefficiency affects both context length and computational requirements, making the models less practical for use. Specifically, the Mistral tokenizer has a compression rate of 5.81 tokens per Hebrew word, which averages approximately one token per character.

\begin{figure}[ht]
    \centering
    \includegraphics[width=0.7\linewidth]{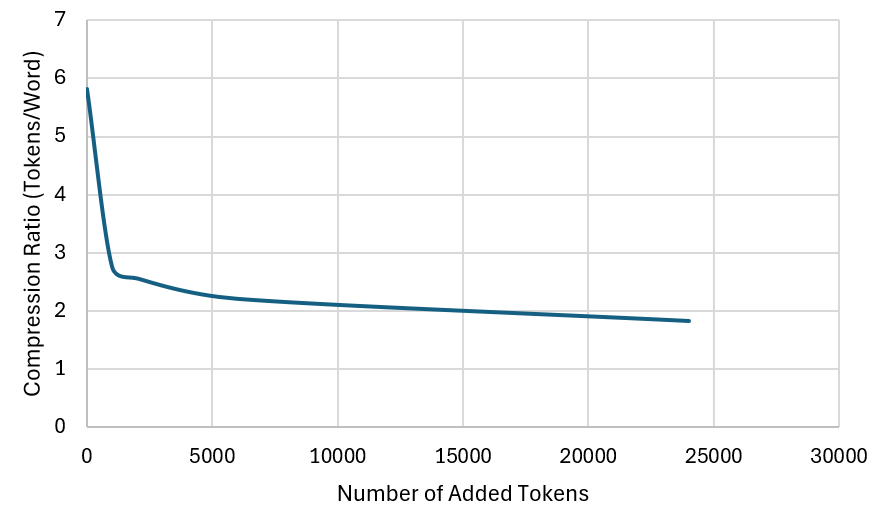}
    \caption{This figure illustrates the relationship between the number of added tokens and the compression ratio in Hebrew (tokens per word).}
    \label{fig:compression_ratio}
\end{figure}

To overcome this, we take inspiration from the works of \citet{touvron2023llama,csaki2023efficiently}, and extend the Mistral tokenizer with 1,000 Hebrew-specific tokens. As shown in Figure \ref{fig:compression_ratio}, adding just 1,000 tokens more than halves the compression rate, after which the gains from adding more tokens diminish significantly.

Additionally, we heed the warning from \citet{zhu2023extrapolating} that extending the tokenizer can degrade model performance, and perform a dedicated training procedure to mitigate this issue. The tuning of the added parameters is split into two stages:

1. \textbf{Embedding Distillation}: As noted by \citet{zhu2023extrapolating}, models perform well when continuing training without extending the tokenizer indicating that there is knowledge in the existing embeddings. Therefore, we aim to distill the existing embeddings in the base model into the newly added tokens. We sample a corpus of approximately 500,000 sentences in Hebrew and train the model to minimize the following loss:

\[ 
\mathcal{L} = \left\| h_{\text{old}} - h_{\text{new}} \right\|_2 
\]

Where \(h_{\text{old}}\) denotes the last hidden state (as in, the last hidden state of the last token) of the output from the model when encoding a sentence using the Mistral tokenizer, and \(h_{\text{new}}\) is the last hidden state of the output from the model when encoding the same sentence using our extended tokenizer. During this training process, we freeze all the weights from the model except the 1,000 added embeddings on the embedding layer. We ran one epoch on this data, using an initial learning rate of \(5e-6\).

2. \textbf{LM-Head Calibration}: Extending the tokenizer adds parameters to both the embedding layer and the final LM Head layer. After tuning the embedding layer parameters, the parameters added to the LM Head layer remain initialized at random. In order to efficiently calibrate those parameters to align with the rest of the trained model, we perform another stage of training where we freeze all model weights except the embedding and LM head layers allowing them to train without affecting the rest of the model. We sample a corpus of 100,000 documents and train the model using the regular next-token prediction objective. We ran one epoch on this data, using an initial learning rate of \(5e-6\).

We evaluated the model after these two stages, and the results are shown in the evaluation section in Figure \ref{fig:leaderboard}. The model is titled \texttt{dicta-il/dictalm2.0-untrained} in the chart, and as can be seen this process alone significantly improves the model's ability to process Hebrew text.\footnote{A similar approach was used by \citet{kim2024efficienteffectivevocabularyexpansion} in their work on efficient and effective vocabulary expansion. Since we had already released our model to the public upon discovering this paper, we did not directly compare the methods.}

\section{Continuous Pre-training}
\label{sec:continue-pretraining}

With the prepared model from Section \ref{sec:model-setup} and the data from Section \ref{sec:pretraining-data}, we began the main training session. This process, called continuous pre-training, involves continuing to pre-train the model on the general next-token-prediction objective using large amounts of unsupervised data.

We conducted the pre-training on a compute cluster of 48 Intel Gaudi-2 chips, using the Optimum-Habana codebase\footnote{\url{https://github.com/huggingface/optimum-habana}} with DeepSpeed Zero-2 \cite{rajbhandari2020zero}. The total training time for one epoch on the data was 15 days. Detailed training hyperparameters are listed in Table \ref{tab:training-hyp}.

\begin{table}[h]
    \centering
    \begin{tabular}{ll}
        \hline
        Global batch size & 384 \\
        Micro batch size & 8 \\
        Sequence length & 4096 \\
        Tokens per global batch & \textasciitilde1.5M \\
        Initial learning rate & 5e-6 \\
        Weight decay & 0.1 \\
        Warmup & 1000 steps \\
        Schedule type & Cosine \\
        \hline
    \end{tabular}
    \caption{Training hyperparameters}
    \label{tab:training-hyp}
\end{table}

\begin{figure}[h]
    \centering
    \includegraphics[width=0.7\linewidth]{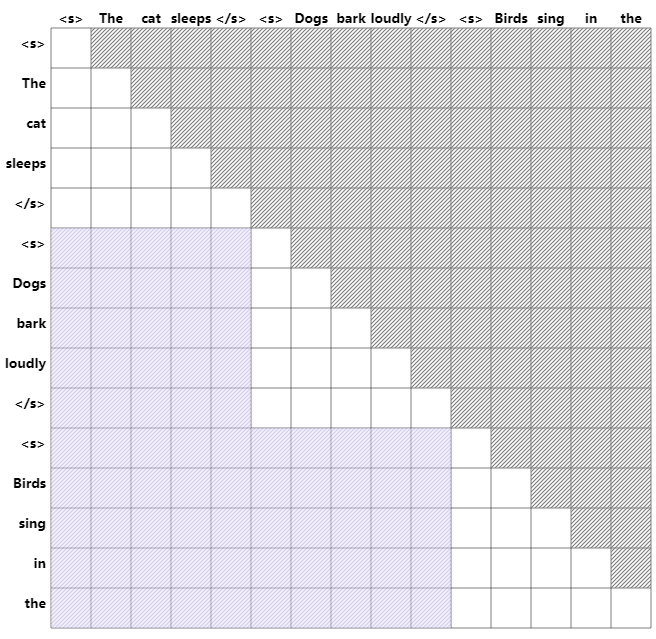}
    \caption{This graph depicts the changes applied to the regular causal mask (dark gray) for our document-attention causal mask (light purple), ensuring tokens from separate documents are masked to restrict cross-document attention.}
    \label{fig:doc-attn}
\end{figure}

To maximize the utilization of available compute power, we pre-processed the data into training sequences using the packing method \cite{li2021efficient}. Each document was wrapped with a BOS (\texttt{<s>}) and EOS (\texttt{</s>}) token. To avoid cross-document attention during training, we employed a document-attention causal mask, ensuring each document attended only to itself (see Figure \ref{fig:doc-attn}).

\begin{figure}[h]
    \centering
    \includegraphics[width=1\linewidth]{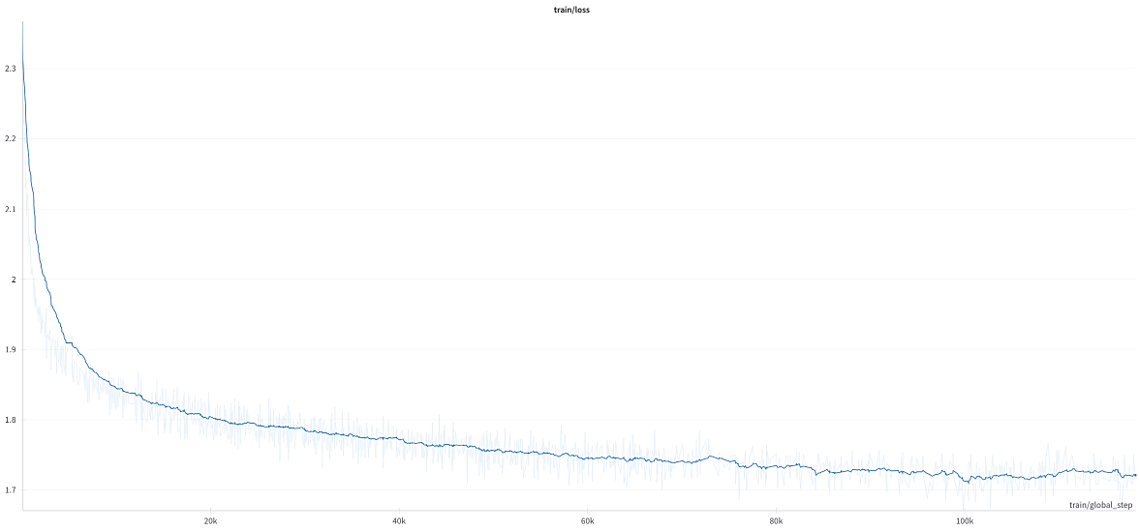}
    \caption{This graph depicts the loss value during the continuous pre-training stage.}
    \label{fig:train-loss}
\end{figure}

The training loss graph is shown in Figure \ref{fig:train-loss}. The loss starts at around 2.5-3, thanks to the initial training done during the tokenizer extension stage, and continues to decrease throughout the epoch. We selected the final checkpoint by reviewing the 10 checkpoints with the lowest loss within the last 10 thousand steps, and of those we chose the checkpoint with the highest evaluation scores. The resulting model, named \texttt{DictaLM2.0}, is freely accessible and available for download on HuggingFace\footnote{\url{https://huggingface.co/dicta-il/dictalm2.0}}.

\section{Instruct Tuning}

Following the creation of our \texttt{DictaLM2.0} model via continuous pre-training, we produced a second model - a chat model - by transitioning to instruct tuning. Instruct tuning involves fine-tuning the model with a curated dataset containing instructions and corresponding responses, enabling the model to better understand and execute tasks based on explicit human instructions. This process enhances the model's ability to follow complex commands and provide more accurate and contextually relevant outputs. Leveraging transfer learning principles, instruct tuning uses the knowledge acquired during continuous pre-training to efficiently adapt to task-specific data and improve performance on downstream applications.

For the process, we followed the methodology presented by \citet{tunstall2023zephyrdirectdistillationlm} used to train their Zephyr-7B-beta model. That model demonstrated impressive performance, and its detailed documentation allowed us to replicate their work on our model. All of their code, models, data, and tutorials are available on GitHub.\footnote{\url{https://github.com/huggingface/alignment-handbook/tree/main/recipes/zephyr-7b-beta}}

\subsection{Supervised Fine-tuning (SFT)}

The first stage of instruct tuning is supervised fine-tuning (SFT). SFT involves training the model on labeled datasets with explicitly defined dialogues. This phase is crucial to develop the model's capability to generate accurate responses and to adhere to desired behavior patterns in real-world applications.

We started by collecting the training dataset for SFT. To train the model to follow instructions in both English and Hebrew, we curated a dataset containing instructions in both languages.

For the English corpus, we used the UltraChat\_200k\footnote{\url{https://huggingface.co/datasets/HuggingFaceH4/ultrachat_200k}} dataset provided by \citet{tunstall2023zephyrdirectdistillationlm}, a filtered version of the original UltraChat dataset \cite{ding2023enhancing}  containing 1.4 million dialogues generated by ChatGPT. Additionally, we sampled 100,000 dialogues from the OpenHermes 2.5 dataset \cite{OpenHermes2.5}, following the success of the Noun-Hermes models.

Collecting the Hebrew instruction corpus was more challenging due to the lack of available instruction datasets in Hebrew. During our evaluation of the DictaLM2.0 base model (detailed in Section \ref{sec:continue-pretraining}), we determined that it is capable of producing high-quality translations between English and Hebrew (see evaluations in Sections \ref{sec:base-eval} and \ref{sec:human-eval}). We leveraged this capability to create our Hebrew instruction dataset, by using our base model to translate the 100,000 dialogues we sampled from the OpenHermes 2.5 dataset and 120,000 dialogues sampled from the UltraChat\_200k dataset from English into Hebrew. Additionally, we included 2,000 high-quality Hebrew dialogues with complex reasoning tasks, which we curated manually in-house.

We also included 25,000 synthetically-created dialogues of translations between English and Hebrew, using approximatley 20 different templates to convert translation pairs from the CCMatrix corpus into dialogues.

SFT Training was conducted on a DGXA100 server with 8x40GB A100-SXM GPUs, using the \texttt{SFTTrainer} from the TRL package \cite{vonwerra2022trl} and DeepSpeed Zero-3 \cite{rajbhandari2020zero}. We trained with LoRA adapters \cite{hu2021loralowrankadaptationlarge}, with $rank=64$ and $\alpha=16$, applied to all linear layers except the embedding layers. We used sequences of 4096 tokens combined with the packing method, a batch size of 64, and trained for a total of 3 epochs.

\subsection{Direct Preference Optimization (DPO)}

The next stage in our model refinement process was Direct Preference Optimization (DPO)  \cite{rafailov2023directpreferenceoptimizationlanguage}. DPO focuses on optimizing the model based on user preferences and feedback, enhancing its ability to generate responses that are both accurate and aligned with user expectations. By integrating real-world feedback into the training loop, DPO helps fine-tune the model's behavior, making it more adaptable and responsive to nuanced user needs. This phase ensures the model not only follows instructions but also evolves to provide contextually appropriate and user-centric outputs.

We followed the methodology of \citet{tunstall2023zephyrdirectdistillationlm} and used the UltraFeedback\_binarized dataset, a preprocessed version of the UltraFeedback dataset \cite{cui2023ultrafeedback} prepared for DPO training. As this dataset includes only English pairs, we translated the entire dataset into Hebrew using the DictaLM2.0 base model, and then combined it with the English data.

During our experiments, we noticed that the model sometimes switched languages within a conversation. To address this, we added a corpus of conversation pairs where the accepted and rejected answers were identical except for language, ensuring the model responded in the same language as the prompt, and penalizing it if it didn't. We created this dataset by sampling entries from the SFT dataset in both English and Hebrew, making up approximately 10\% of the final DPO corpus.
 
DPO training was conducted on a DGXA100 server with 8x40GB A100-SXM GPUs, using the \texttt{DPOTrainer} from the TRL package \cite{vonwerra2022trl} and DeepSpeed Zero-3 \cite{rajbhandari2020zero}. We trained with LoRA adapterss \cite{hu2021loralowrankadaptationlarge}, with $rank=64$ and $\alpha=16$, applied to all linear layers except the embedding layers. We set beta to 0.01 (following the experiments done by \citet{tunstall2023zephyrdirectdistillationlm}) and trained for one epoch with a learning rate of \(5e-7\) and a warmup ratio of \(10\%\).

We named the final model produced from the SFT and DPO training \texttt{DictaLM2.0-instruct}, and it is freely accessible and available for download on HuggingFace.\footnote{\url{https://huggingface.co/dicta-il/dictalm2.0-instruct}}
 
\section{Evaluation}

Evaluation is a critical component of model development, providing essential metrics to assess quality and performance. Despite the advancements in developing \texttt{DictaLM2.0} and \texttt{DictaLM2.0-Instruct}, the initial scarcity of relevant evaluation sets for Hebrew LLMs posed a challenge in fully validating their effectiveness, reliability, and robustness in various applications. To address this gap, we created comprehensive evaluation methods and datasets tailored specifically for benchmarking Hebrew LLMs.

To thoroughly assess our models, we developed an evaluation framework that encompasses three primary categories, detailed below:

\subsection{Automatic Evaluation}
\label{sec:base-eval}
Our first method of evaluation employs automatic assessment of the base model's responses to tasks using few-shot learning with a human-curated test set. Automatic evaluation is essential for providing scalable, consistent, and objective metrics to measure model performance across diverse tasks. Evaluating generative language models poses unique challenges due to their training objective of next-token prediction, which complicates reliable assessment on specific downstream applications. However, as demonstrated by \citet{brown2020languagemodelsfewshotlearners}, language models possess the ability for in-context learning (ICL), allowing them to perform tasks with minimal examples (few-shot prompts). This capability facilitates the creation of few-shot prompts that effectively guide the model to comprehend the task and generate appropriate responses. To rigorously evaluate the model's capabilities, we curated four datasets designed to test different aspects of the model’s performance:

\begin{itemize}
    \item \textbf{Hebrew Question Answering}: This task evaluates a model's ability to understand and process information presented in Hebrew, focusing on comprehension and the accurate retrieval of answers based on context. It checks the model's grasp of Hebrew syntax and semantics through direct question-and-answer formats.
        \begin{itemize}
            \item \textbf{Source}: The test subset of the HeQ dataset \cite{cohen-etal-2023-heq}, which contains 1,436 entries. 
            \item \textbf{Scoring}: Results are scored automatically using the \texttt{tlnls} scoring method proposed by \citet{cohen-etal-2023-heq}, which accounts for the linguistic properties of Hebrew language.
            \item \textbf{Few-Shot Format}: Each context paragraph is followed by three questions and answers, and finally the desired question unanswered.
        \end{itemize}
    \item \textbf{Sentiment Analysis}: This task tests the model's ability to detect and interpret sentiments in Hebrew text. It assesses the model's capability to classify statements accurately as positive, negative, or neutral based on linguistic cues.
        \begin{itemize}
            \item \textbf{Source}: Hebrew Sentiment \cite{hebrew_sentiment_dataset_2024} - a Sentiment-Analysis Dataset in Hebrew. Because many of the sentences in the dataset are dubious or ambiguous, we employed a professional linguist who selected a set of 3,000 fairly unambiguous cases from the dataset. 
            \item \textbf{Scoring}: The model must predict one of the classes (Positive, Negative, Neutral), and accuracy is computed automatically.
            \item \textbf{Few-Shot Format}:Each prompt includes the same nine few-shot examples, three from each category, randomly shuffled.
        \end{itemize}
    \item \textbf{Winograd Schema Challenge}: This task measures the model’s understanding of pronoun resolution and contextual ambiguity in Hebrew. It tests the model’s ability to use logical reasoning and general world knowledge to disambiguate pronouns correctly in complex sentences.
        \begin{itemize}
            \item \textbf{Source}: A translation of the Winograd Schema Challenge to the Hebrew language \cite{shwartz2021wsc}.
            \item \textbf{Scoring}: The model must choose which of two entities the question refers to, with the options presented in the prompt. Accuracy is computed automatically. 
            \item \textbf{Few-Shot Format}: Each prompt includes five few-shot examples, with the input sentence, question, possible answers, and expected answer.
        \end{itemize}
    \item \textbf{Translation}: This task assesses the model's proficiency in translating between English and Hebrew. It evaluates the linguistic accuracy, fluency, and the ability to preserve meaning across languages, highlighting the model’s capability in bilingual translation tasks.
        \begin{itemize}
            \item \textbf{Source}: NeuLabs-TedTalks aligned translation corpus \cite{tiedemann-2012-parallel}. We sampled 1,000 sentences confirmed not to appear in CCMatrix (since the CCMatrix was included in our training corpus). We tested the ability of the model to translate in both directions. 
            \item \textbf{Scoring}: Translations are compared to the gold translation from the corpus using the BLEU score \cite{papineni-etal-2002-bleu}, the standard scoring method for machine translation.
            \item \textbf{Few-Shot Format}: Each prompt includes three few-shot examples of an English sentence and its Hebrew equivalent (or the converse, for cases of Hebrew to English translation).
        \end{itemize}
\end{itemize}

\begin{figure}[h]
    \centering
    \includegraphics[width=1\linewidth]{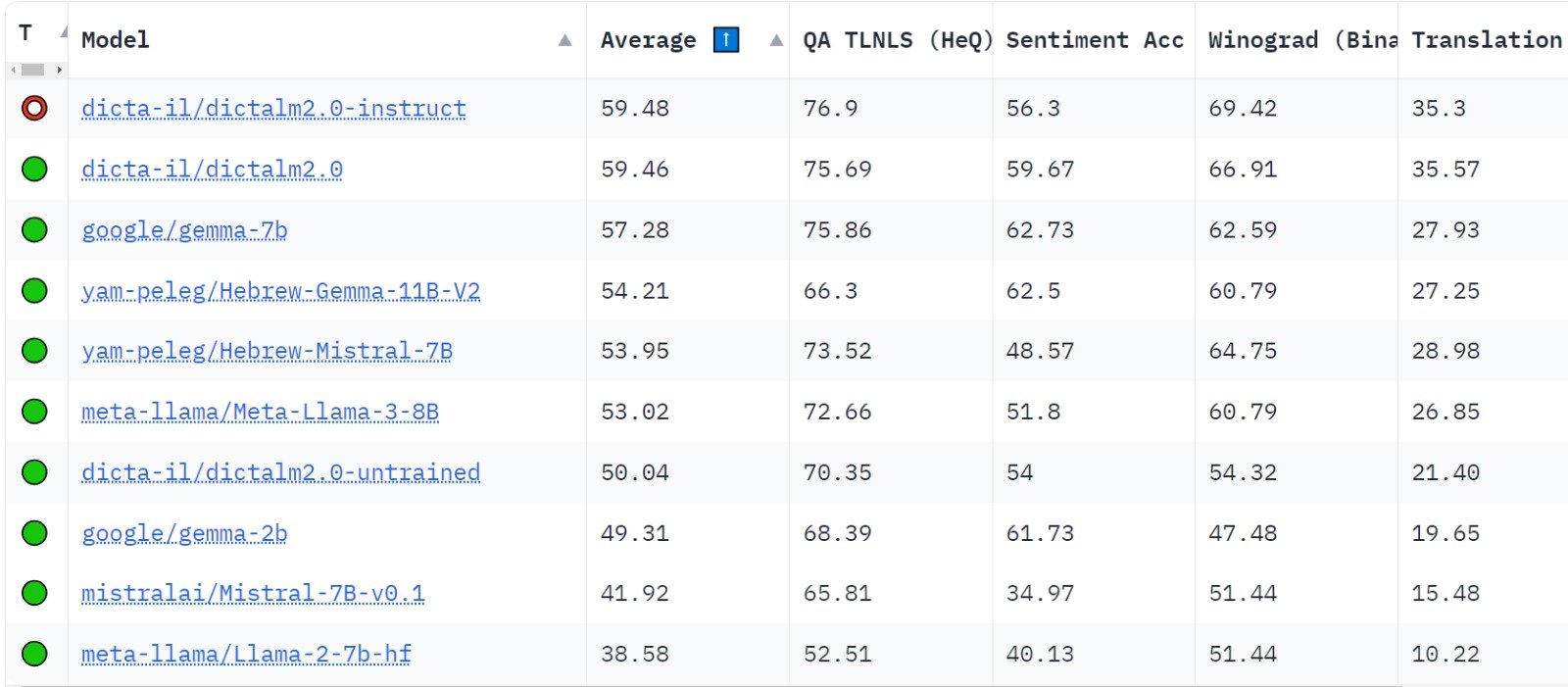}
    \caption{Comparison of evaluation results between our model and other base models on Hebrew few-shot tasks.}
    \label{fig:leaderboard}
\end{figure}

We present the results comparing our two models (DictaLM2.0 and DictaLM2.0-Instruct) to other SOTA base models in the 7B parameter range, as well as to other models that claim to perform well on Hebrew. The results are displayed in Figure \ref{fig:leaderboard}. Our models produce SOTA results on most of these tasks, particularly regarding the Winograd and Translation benchmarks.

\subsection{Human Evaluation}
\label{sec:human-eval}

The second method of evaluation employed was human judgment, also known as "Human as a Judge." We conducted a blind test in which human evaluators were presented with translations from our base model alongside translations from Google Translate. The base model was provided with a few-shot prompt containing three translation pairs preceding each input sentence to indicate the nature of the task. To ensure unbiased assessments, the translations were anonymized, and evaluators were asked to choose the better translation without knowing which system produced it. This method offers valuable insight into the comparative quality of our model's translations.

\begin{figure}[h]
    \centering
    \includegraphics[width=0.7\linewidth]{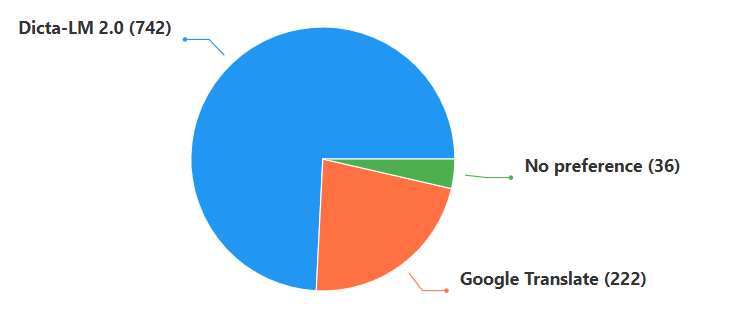}
    \caption{Human evaluation results from a blind test comparing translations from our model and Google Translate.}
    \label{fig:human-judge-results}
\end{figure}

We conducted the test on 1,000 sentences translated from English to Hebrew. The results, presented in Figure \ref{fig:human-judge-results}, show a strong preference for the translations produced by our model over those from Google Translate. These successful results led us to set up a translation site for the public to use free of charge, for translating between English and Hebrew, in both directions. The site is available online.\footnote{\url{https://translate.dicta.org.il}}

\subsection{LLM-based Evaluation}

For the final evaluation method, we aim to assess a Hebrew chat model's ability to understand and follow instructions, as well as the quality of the generated content. In order to do so, we utilize a summarization test set. We collected a corpus of 75 Hebrew news documents with human-curated summaries. We then instruct each of the chat models to generate summaries for each of the documents. Summaries are generated with a temperature of 0.7 using a generic prompt describing the task of summarizing the document in a short 4-5 sentence paragraph. We use GPT-4 as a judge to score the summaries.

For each model, the original documents and the generated summaries were presented to GPT-4 for evaluation. The human-curated summaries were scored as well in order to provide a baseline to compare the model summaries to. We used the scoring prompts described in OpenAI's cookbook,\footnote{\url{https://cookbook.openai.com/examples/evaluation/how_to_eval_abstractive_summarization}} adjusting them to a 1-10 scale. GPT-4 scored each summary on four measures:

\begin{itemize}
    \item \textbf{Relevance} - Selecting important content from the original source.
    \item \textbf{Coherence} - The collective quality of all sentences.
    \item \textbf{Consistency} - Factual alignment between the summary and the original content.
    \item \textbf{Fluency} - Quality of the summary in terms of grammar, spelling, punctuation, word choice, and sentence structure.
\end{itemize}

The results are presented in Table \ref{tab:evaluation-scores}. We compare the performance of our chat model to popular proprietary models from OpenAI and Anthropic, as well as to popular open-source instruct-tuned models like Gemma-7B-it \cite{gemma_2024} and LLaMA-3-8B-Instruct \cite{llama3modelcard}. We evaluate our chat model both after SFT instruct training and after the DPO training. The DPO training significantly improves the model's performance. While our model is not yet competitive with proprietary models, it provides a high-quality open-source alternative for those looking to fine-tune a model for specific needs.

\begin{table}[h]
    \centering
    \begin{tabular}{lcccc}
        \hline
        Model & Relevance & Coherence & Consistency & Fluency \\
        \hline
        google/gemma-7b-it & 6.29 & 5.37 & 7.25 & 7.19 \\
        meta-llama/Llama-3-8B-Instruct & 7.7 & 6.96 & 8.53 & 8.14 \\
        GPT 3.5 Turbo & 7.55 & 6.76 & 8.4 & 8.12 \\
        DictaLM2.0 Instruct (SFT) & 7.45 & 6.85 & 7.93 & 8.17 \\
        DictaLM2.0 Instruct (SFT + DPO) & 8.1 & 7.45 & 8.34 & 8.54 \\
        Claude 3 Haiku & \textbf{8.73} & 7.95 & \textbf{9.67} & \textbf{8.69} \\
        GPT 4 11-06 Preview & 8.65 & \textbf{7.96} & 9.48 & \textbf{8.69} \\
        Human & 8.33 & 7.66 & 9.42 & 8.59 \\
        \hline
    \end{tabular}
    \caption{Evaluation of instruct-tuned models on the LLM-evaluated summarization task. The DictaLM2.0-Instruct model is presented both after SFT and after DPO to demonstrate the importance and efficacy of the DPO training.}
    \label{tab:evaluation-scores}
\end{table}

\subsection{Hebrew Open LLM Leaderboard}

We compiled all the benchmarks from Section \ref{sec:base-eval} and created a Hebrew Open LLM Leaderboard, which is now publicly available. This leaderboard serves as a call to action for responsible research and development in Hebrew LLMs. For the first time, it offers a consistent benchmark setup that ensures reliable results, making it easier for everyone to evaluate and compare models effectively. By providing this resource, we aim to foster innovation and collaboration within the community. The leaderboard can be accessed here: \url{https://huggingface.co/spaces/hebrew-llm-leaderboard/leaderboard}.

\section{Conclusion}

The development and release of \texttt{DictaLM2.0} and \texttt{DictaLM2.0-Instruct} represent major advancements in the field of Hebrew natural language processing. By addressing the unique challenges of Hebrew and other low resource languges, we have optimized these generative models for high performance in Hebrew. Through rigorous evaluation, both automated and human-judged, our models have demonstrated state-of-the-art performance across various NLP tasks such as question answering, sentiment analysis, and machine translation.

Furthermore, we introduced a comprehensive benchmark suite and an open Hebrew LLM leaderboard, providing essential resources for future research and evaluation in Hebrew NLP. This work not only sets new standards for Hebrew language models but also offers methodologies that can be adapted to other low-resource languages, contributing to the broader field of multilingual NLP.

In conclusion, \texttt{DictaLM2.0} and \texttt{DictaLM2.0-Instruct} represent robust, open-source solutions that pave the way for further innovations in Hebrew NLP. By fostering ongoing research and development within the community, these models and resources support the advancement of NLP capabilities for low-resource languages, enhancing the inclusivity and diversity of language technologies.

\section*{Acknowledgements}

We would like to express our thanks to Intel NLP Labs Israel: Peter Izsak, Daniel Fleischer, Moshe Berchansky and Moshe Wasserblat, for their expertise and support in training the model on the Intel Gaudi systems. 

We would also like to thank Avner Algom and the Israeli Association of Human Language Technologies (IAHLT) for making the connection with Intel and thereby paving the way for the creation of this model.

We would also like to thank Tal Geva and the DDR\&D IMOD / The Israeli National Program for NLP in Hebrew and Arabic for their invaluable contributions without which this project wouldn't have been possible.

\bibliography{anthology,custom}
\bibliographystyle{acl_natbib}

\end{document}